\documentclass[10pt,twocolumn,letterpaper]{article}

\usepackage[pagenumbers]{wacv} 

\definecolor{wacvblue}{rgb}{0.21,0.49,0.74}
\usepackage[pagebackref,breaklinks,colorlinks,allcolors=wacvblue]{hyperref}

\def\wacvPaperID{3196} 
\def\confName{WACV}
\def\confYear{2026}

\title{Explainable Human-in-the-Loop Segmentation via Critic Feedback Signals}

\author{Pouya Shaeri\\
Arizona State University\\
Tempe, AZ 85281\\
{\tt\small pshaeri@asu.edu}
\and
Ryan T. Woo\\
Arizona State University\\
Tempe, AZ 85281\\
{\tt\small rtwoo@asu.edu}
\and
Yasaman Mohammadpour\\
Arizona State University\\
Tempe, AZ 85281\\
{\tt\small ymoham15@asu.edu}
\and
Ariane Middel\\
Arizona State University\\
Tempe, AZ 85281\\
{\tt\small amiddel@asu.edu}
}

\begin{document}
\maketitle
\begin{abstract}
Segmentation models achieve high accuracy on benchmarks but often fail in real-world domains by relying on spurious correlations instead of true object boundaries. We propose a human-in-the-loop interactive framework that enables interventional learning through targeted human corrections of segmentation outputs. Our approach treats human corrections as interventional signals that show when reliance on superficial features (e.g., color or texture) is inappropriate. The system learns from these interventions by propagating correction-informed edits across visually similar images, effectively steering the model toward robust, semantically meaningful features rather than dataset-specific artifacts. Unlike traditional annotation approaches that simply provide more training data, our method explicitly identifies when and why the model fails and then systematically corrects these failure modes across the entire dataset. Through iterative human feedback, the system develops increasingly robust representations that generalize better to novel domains and resist artifactual correlations. We demonstrate that our framework improves segmentation accuracy by up to 9 mIoU points (12-15\% relative improvement) on challenging cubemap data and yields 3-4$\times$ reductions in annotation effort compared to standard retraining, while maintaining competitive performance on benchmark datasets. This work provides a practical framework for researchers and practitioners seeking to build segmentation systems that are accurate, robust to dataset biases, data-efficient, and adaptable to real-world domains such as urban climate monitoring and autonomous driving.

\end{abstract}

\section{Introduction}
\label{sec:intro}

Semantic segmentation is a cornerstone of computer vision, enabling dense prediction tasks such as autonomous driving, medical diagnostics, urban scene analysis, and environmental monitoring. The past decade has seen rapid progress with deep learning models such as DeepLab~\cite{chen2017deeplab}, U-Net~\cite{ronneberger2015u}, SegFormer~\cite{xie2021segformer}, and Mask2Former~\cite{cheng2022masked}, which consistently achieve state-of-the-art performance on standard benchmarks including Cityscapes~\cite{cordts2016cityscapes}, ADE20K~\cite{zhou2017scene}, and COCO-Stuff~\cite{caesar2018coco}. However, despite these advances, segmentation models continue to underperform in real-world deployments where test distributions diverge from the curated benchmarks on which they were trained.

The brittleness of deep segmentation models is increasingly recognized as a consequence of their tendency to exploit superficial correlations rather than learn interventionally relevant features~\cite{geirhos2020shortcut}. For example, models may classify all blue regions as ``sky'' even when those pixels correspond to buildings or vehicles, or they may rely on texture heuristics that misclassify natural rock formations as man-made structures. These shortcut strategies yield high accuracy on training distributions but fail under domain shift, occlusion, or rare contexts~\cite{hendrycks2019benchmarking}. This fragility is particularly problematic in safety-critical applications such as autonomous navigation~\cite{sun2020scalability}, medical imaging~\cite{litjens2017survey}, and climate science~\cite{shaeri2025multimodalphysicsinformedneuralnetwork,alkhaled2024webmrt}, where dataset biases can undermine reliability and trust~\cite{shaeri2025sentiment}. A wide range of approaches have been proposed to mitigate this problem. Training-based methods include extensive data augmentation~\cite{chen2020simple}, adversarial training~\cite{madry2017towards}, domain adaptation~\cite{ganin2015unsupervised}, and interventional representation learning~\cite{scholkopf2021toward}. While effective in controlled experiments, these strategies often require costly retraining whenever new failure modes are discovered. Moreover, they may not generalize to unforeseen correlations that were not anticipated during training. On the other hand, human-in-the-loop approaches focus on leveraging human expertise to guide model improvement. Examples include active learning for pixel annotations~\cite{cai2021revisiting}, weakly supervised labeling~\cite{papandreou2015weakly}, or interactive refinement tools such as GrabCut~\cite{rother2004grabcut} and SAM-based prompting~\cite{kirillov2023segment}. 

\begin{figure*}[t]
    \centering
    \includegraphics[width=\textwidth]{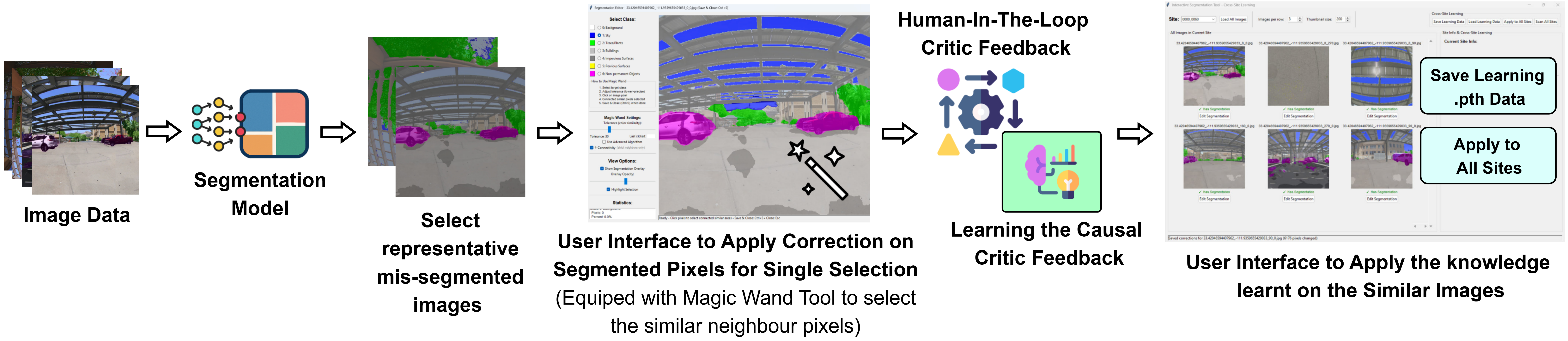}
    \caption{Overview of our human-in-the-loop segmentation with correction propagation.}
    \label{fig:framework}
\end{figure*}

In this work, we propose a new perspective as illustrated in Figure~\ref{fig:framework}: treating human corrections not merely as additional labels but as interventional signals that provide counterfactual evidence about model behavior. Inspired by interventional reasoning principles~\cite{pearl2018book}, we view a human correction as an interventional supervision signal:
\[
\text{segmentation}[R] \leftarrow y^{*},
\]
where $R$ is the corrected region and $y^{*}$ the user-specified class. This reframing emphasizes that corrections go beyond passive labels by explicitly overriding the model’s correlation-driven prediction. This explicitly signals that, under the same visual input, the model’s reliance on a spurious correlation (e.g., “green pixels = vegetation”) is invalid, and the semantically correct classification is different. Each correction thus provides valuable interventional data that the model cannot extract from passive training alone. Building on this idea, we design a human-in-the-loop interactive framework that transforms segmentation error correction into a process of interventional learning. The framework integrates three intertwined mechanisms: a Critic Interface, which provides a visual editing tool that allows humans to not only fix segmentation errors but also give targeted feedback on why the prediction was wrong; Counterfactual Data Generation, where each correction produces counterfactual pairs contrasting the original correlation-driven prediction with the interventionally corrected segmentation; and Feedback Propagation, which extends corrections across visually similar images by effectively asking, “If intervention was necessary for image A, should the same correction apply to image B?” This mechanism broadens propagation across datasets beyond single images to entire datasets.

Unlike prior interactive segmentation systems that passively incorporate human annotations, our framework treats human expertise as direct supervision signals. This distinction is crucial. By explicitly identifying and breaking shortcut strategies (color heuristics, texture biases, contextual assumptions), our system guides the model toward more robust, semantically meaningful representations. In contrast, simply adding more data through annotation often reinforces non-robust cues if the underlying bias remains unaddressed.

Furthermore, interventional framing aligns naturally with robustness evaluation: rather than measuring raw pixel accuracy alone, we assess spurious correlation resistance, capturing model performance when superficial correlations are violated or inverted; cross-domain generalization, reflecting the transferability of corrections to datasets with different correlation structures; and interventional feature learning, which involves both qualitative and quantitative analysis of the features (such as shapes, spatial relationships, and semantic context) the model learns to rely on after human interventions. This evaluation lens enables us to move beyond simple error fixing toward measuring interventional robustness, a growing focus in the vision community~\cite{scholkopf2021toward,mitrovic2020representation}. To summarize, this paper makes several key contributions: we propose a human-in-the-loop interventional learning framework that treats corrections as explicit interventions, providing counterfactual-style signals to break artificial correlations in segmentation models; we design a model-agnostic critic interface that supports interactive segmentation editing and visualizes why predictions fail, enabling reasoning about model errors; we introduce a Feedback Propagation mechanism that generalizes corrections across visually similar images, effectively scaling correction propagation across datasets with limited human effort; We provide a comprehensive evaluation on benchmark datasets (Cityscapes, ADE20K) and domain-specific cubemap imagery, showing that our method reduces segmentation errors, improves cross-domain robustness, and achieves performance on par with human annotations while outperforming retraining baselines. We position our work as among the first to connect interventional representation learning with interactive segmentation, contributing both methodological insights and practical tools for robust deployment.

\section{Related Work}
\label{sec:related-work}

Research on semantic segmentation, human-in-the-loop learning, explainability, and causal inference has evolved rapidly in recent years. In this section, we review related efforts and situate our work at the intersection of interactive segmentation, human-in-the-loop machine learning, explainable AI, and causal learning in computer vision.

\subsection{Interactive Segmentation}

Interactive segmentation has been a practical solution to reduce annotation costs while maintaining high-quality masks. Early systems such as Graph Cuts~\cite{boykov2001interactive} and Random Walks~\cite{grady2006random} relied on user-provided scribbles or bounding boxes to refine object boundaries. Later works incorporated geodesic distances~\cite{gulshan2010geodesic} and region-growing strategies to reduce annotation effort. With the advent of deep learning, neural networks have been increasingly integrated into interactive frameworks, enabling rapid propagation of user corrections across an image~\cite{xu2016deep, majumder2019content}. Recent advances aim to minimize the number of user interactions. Models such as F-BRS~\cite{sofiiuk2020f} introduced fast backpropagation refinements to speed up corrections, while methods like RITM~\cite{sofiiuk2022reviving} emphasize iterative minimal interactions. More recently, the Segment Anything Model (SAM)~\cite{kirillov2023segment} has demonstrated zero-shot segmentation capabilities through point and box prompts, making interactive segmentation scalable to a wide range of domains. However, despite their success, these systems largely interpret human input as additional supervision (points, masks, or bounding boxes) without incorporating human reasoning about why the model failed. As a result, systematic biases and spurious patterns remain unaddressed. Our work complements and extends this line of research by reinterpreting human corrections as interventional signals. Instead of treating user input as merely additional labels, we use them as counterfactual-style feedback that highlights and helps break model reliance on shortcuts. This moves interactive segmentation beyond efficiency improvements toward improved robustness.

\subsection{Human-in-the-Loop Machine Learning}

Human-in-the-loop (HITL) methods have a long history in machine learning~\cite{decost2017computer,shaeri2025mnist}. In vision, HITL frameworks are often applied in active learning~\cite{settles2009active}, where the system queries humans for labels on uncertain or representative examples. Examples include uncertainty sampling for semantic segmentation~\cite{kasarla2019region}, query-by-committee approaches~\cite{beluch2018power}, and Bayesian active learning~\cite{gal2017deep}. These methods reduce labeling cost but still assume that human input is limited to labeling ambiguous instances.

Beyond active learning, researchers have explored richer forms of interaction. Preference-based reinforcement learning~\cite{christiano2017deep} leverages human judgments to align models with subjective criteria, while interactive debugging systems~\cite{kulesza2015principles} allow users to iteratively refine models based on interpretable errors. In computer vision, HITL tools have been designed for annotation refinement~\cite{lin2016scribblesup}, weak supervision~\cite{papandreou2015weakly}, and dataset curation~\cite{mosqueira2023human}.

Our framework differs by explicitly embedding causal reasoning into the loop. Instead of humans serving as annotators or preference providers, they act as critics who not only provide corrections but also indicate why predictions are wrong (e.g., reliance on texture instead of object boundaries). This distinction enables corrections that propagate beyond individual examples, improving robustness.

\subsection{Explainable AI in Computer Vision}

Explainable AI (XAI) has become central in computer vision, with methods such as gradient-based saliency maps~\cite{simonyan2013deep}, attention visualization in transformers~\cite{dosovitskiy2020image}, and feature attribution techniques like LIME~\cite{ribeiro2016should} and SHAP~\cite{lundberg2017unified}. Pixel-level explanations for segmentation~\cite{fong2019understanding} and counterfactuals~\cite{goyal2019counterfactual} extend interpretability, though these approaches remain largely \emph{passive}. Interactive explainability tools such as GAMUT~\cite{hohman2019gamut} and Explainer Studio~\cite{spinner2019explainer} provide visual exploration, yet rarely allow human feedback to update models. Our framework addresses this gap by integrating explanation with critic feedback, enabling a critic interface where users both interpret errors and directly refine model reasoning~\cite{arrieta2020explainable}.

\subsection{Causal Inference and Robustness in Vision}

Causality has emerged as a key principle for robust machine learning~\cite{pearl2018book,scholkopf2021toward}, with applications in computer vision ranging from domain adaptation~\cite{zhang2021causal} and visual question answering~\cite{beigi2024can} to bias mitigation~\cite{wang2020towards} and robust representation learning~\cite{mitrovic2020representation}. The goal is to disentangle shortcut correlations from causal factors, enabling generalizable representations. Recent efforts incorporate weak supervision, such as grouping constraints or auxiliary labels, to guide causal feature extraction~\cite{wang2021causal}.

Our work lies in introducing humans into the learning loop by treating their corrections as interventional feedback that complements automated approaches. Human expertise provides direct signals about which correlations are spurious and which features are more semantically relevant, while our propagation mechanism leverages these corrections to scale improvements across datasets. This provides a practical way to incorporate targeted supervision in an interactive setting. Taken together, prior research has advanced interactive segmentation, human-in-the-loop learning, explainable AI, and causal inference as largely separate threads; our work connects these areas through a unified framework with four key innovations. Unlike traditional interactive segmentation, we frame user corrections as critic feedback rather than passive labels. Unlike prior HITL approaches, we enable users to act as critics who not only provide corrections but also highlight dataset biases driving errors. Unlike standard XAI methods, our critic interface extends beyond explanation to enable actionable interventions that update the model in real time. Unlike purely algorithmic robustness methods, we leverage human expertise as direct supervision signals, providing scalable guidance. By combining these threads, we extend human-in-the-loop segmentation from error correction toward systematic robustness, yielding models that are both more reliable and more data-efficient, thereby positioning our work as a novel contribution within the vision community.

\section{Methodology}
\label{sec:methodology}

Our human-in-the-loop interventional segmentation framework integrates state-of-the-art segmentation models with an interactive critic interface and an interventional feedback pipeline\footnote{All implementation details needed to reproduce our experiments are included in the paper. The cubemap data and full source code will be released publicly following the peer review process.}
. The framework consists of three major components: a segmentation backbone, where a model-agnostic engine (SegFormer, Mask2Former, SAM, etc.) produces initial masks and class predictions; an explainable critic interface, implemented as a Tkinter-based interactive editor that visualizes model predictions, highlights failure regions, and enables humans to provide targeted corrections; and an intervention and propagation module, where human corrections are treated as explicit feedback signals that generate counterfactual-style training examples and are propagated to visually similar images, thereby enabling dataset-wide correction propagation with minimal annotation effort as one of the limitations mentioned in~\cite{shaeri2025sentimentsocialsignalsclimate} we address.

The process follows an iterative loop inspired by the interactive loop in ~\cite{shaeri2025mnistgenmodularmniststyledataset}: the backbone predicts segmentations, the critic interface detects and visualizes errors, humans intervene by correcting masks, and these interventions are propagated across the dataset to improve robustness. Crucially, the framework is \emph{model-agnostic}: any backbone can be plugged in through standardized feature and prediction interfaces.

\subsection{Segmentation Backbone}

We support multiple segmentation architectures to demonstrate the generality of our framework:

\begin{itemize}
    \item \textbf{Transformer-based models.} SegFormer (B0–B5 variants) provides efficient hierarchical representations with lightweight decoders~\cite{xie2021segformer}. Developed by unfreezing the last layers and retrain those layers, introduced and done in ~\cite{shaeri2025multimodal}.
    \item \textbf{Mask-based models.} Mask2Former uses a masked attention mechanism for high-quality boundary refinement~\cite{cheng2022masked}.
    \item \textbf{Foundation models.} SAM~\cite{kirillov2023segment} enables prompt-based zero-shot segmentation but lacks mechanisms for correction.
\end{itemize}

In practice, the system invokes \small{\texttt{segment\_image\_adv}}\normalsize, which augments input images with preprocessing (e.g., contrast enhancement for upward-facing fisheye views) and post-processing (morphological cleanup for sky regions). Each backbone outputs pixel-level predictions in a standardized 7-class taxonomy: \emph{sky}, \emph{trees/plants}, \emph{buildings}, \emph{impervious surfaces}, \emph{pervious surfaces}, \emph{non-permanent objects}, and \emph{background}.

\subsection{Explainable Critic Interface}

The critic interface is the central human-facing component implemented in \small{\texttt{SegmentationEditor}}\normalsize. Unlike traditional annotation tools, it emphasizes \emph{why} predictions are wrong rather than just collecting corrected masks.

\subsubsection{Failure Detection}

The interface surfaces regions likely to contain errors based on three complementary criteria: uncertainty detection, where pixels with high entropy across class logits are flagged as unreliable; consistency analysis, where disagreement across ensemble backbones or augmentations reveals systematic brittleness; and feature attribution, where visual saliency maps (e.g., Integrated Gradients~\cite{sundararajan2017axiomatic}) highlight cases where predictions are driven by superficial cues such as color or texture. Together, these signals guide human attention toward regions where interventions yield the greatest corrective value.

\subsubsection{Interactive Editing User Interface}

The magic wand tool is designed to accelerate correction by allowing users to select entire regions with a single click rather than manually outlining them. When a user clicks on a pixel, the tool performs region growing, automatically selecting all connected pixels that fall within a similarity threshold. The threshold (or tolerance) is adjustable: a low tolerance restricts the selection to pixels nearly identical in color or texture to the clicked pixel, while a higher tolerance expands the selection to include a broader range of similar pixels. This makes it possible to quickly capture homogeneous regions such as sky, grass, or building facades. In practice, the tool can leverage both raw image features (e.g., RGB values, texture descriptors) and intermediate feature maps from the segmentation backbone, ensuring that selections align with semantic patterns rather than just low-level pixel values. Users can then refine the selection (expand, shrink, or undo parts) before applying a class reassignment. In this way, the magic wand tool provides a balance between automation and human control, greatly reducing annotation time while keeping the corrections semantically meaningful. Corrected masks are saved in three formats: (1) raw binary files (\texttt{.bin}), (2) indexed PNG maps, and (3) colorized visualizations for qualitative inspection.

\subsubsection{Types of Interventions}

From the codebase, three recurring correction types emerge:

\begin{itemize}
    \item \textbf{Feature suppression.} When a model misclassifies based on superficial cues (e.g., ``all blue = sky''), the correction suppresses reliance on color features in that region.
    \item \textbf{Boundary refinement.} Corrections emphasize object edges and shape cues over texture heuristics.
    \item \textbf{Context reweighting.} Users can override biases from spatial priors (e.g., ``green at top = vegetation’’) by reassigning classes in atypical contexts.
\end{itemize}

\subsubsection{Counterfactual Learning}

For clarity, we denote an input image as $x$ with pixel set $\Omega$. The segmentation backbone is parameterized by $f_\theta$, producing pixel-level predictions $f_\theta(x)_i$ for each $i \in \Omega$. Human corrections are defined over a subset $R \subseteq \Omega$, where corrected labels $y^*_i$ are provided. For propagation across images, $M$ denotes pairs of matched pixels $(i,j)$ identified via similarity search.

Each correction generates a counterfactual training signal:
\[
(x, \hat{y}, y^{*}),
\]
where $x$ is the input image, $\hat{y}$ the backbone prediction, and $y^{*}$ the human-corrected mask. These triples form a dataset of interventions used to refine model parameters. The training objective extends the standard segmentation loss:
\begin{equation}
\mathcal{L}_{total} = \mathcal{L}_{seg} + \lambda_{cf} \mathcal{L}_{cf} + \lambda_{prop} \mathcal{L}_{prop},
\end{equation}
where $\mathcal{L}_{seg}$ is the cross-entropy segmentation loss, $\mathcal{L}_{cf}$ enforces consistency with counterfactual corrections, and $\mathcal{L}_{prop}$ encourages consistency when propagating corrections across visually similar images.

The counterfactual loss is defined as:
\[
\mathcal{L}_{cf} = \frac{1}{|R|}\sum_{i \in R} \ell\big(f_\theta(x)_i, y^*_i\big),
\]
where $R$ is the corrected region. This encourages alignment between predictions and human-provided counterfactuals.

The propagation loss transfers corrections across images:
\[
\mathcal{L}_{prop} = \frac{1}{|M|}\sum_{(i,j) \in M} \ell\big(f_\theta(x^j)_i, y^*_i\big),
\]
where $M$ is the set of pixel correspondences retrieved via similarity search. This ensures that if a correction is valid in one image, similar regions in other images are updated consistently.

\subsection{Similarity-Based Feedback Propagation}

One unique element of our framework is the propagation of corrections across images. Implemented in \texttt{SegmentationLearner}, the system stores corrected region histograms and performs nearest-neighbor search to identify visually similar regions across other sites. For each human correction, the system extracts descriptive features such as color distributions and texture statistics, which capture the visual signature of the corrected region. These features are compared against a global database built from all sites, and the most similar regions are retrieved using efficient nearest-neighbor search. When a match is found, the previously stored correction is automatically transferred, ensuring that if a superficial correlation is broken in one image, the same reasoning can be applied consistently to others. This mechanism transforms a single user edit into a dataset-wide correction signal, reducing redundancy and extending the reach of human expertise. In practice, this propagation reduces manual effort extensively, compared to baseline re-annotation pipelines, while simultaneously supporting large-scale correction propagation.

\subsection{Model-Agnostic Integration}

Our framework is intentionally model-agnostic. For transformer-based backbones (SegFormer), we exploit attention tokens for uncertainty and attribution analysis. For CNN-based models (Mask2Former), intermediate feature maps are exposed for critic visualization. For foundation models (SAM), we use mask embeddings and prompt tokens as hooks for feedback editing.

The Tkinter-based critic interface and propagation mechanism remain constant across backbones. This modularity enables fair comparison of robustness improvements across architectures.

\subsection{Sky-Specific Enhancements}

Given the importance of sky segmentation in urban climate applications, we incorporate specialized preprocessing and post-processing for upward-facing fisheye images:
Contrast-limited adaptive histogram equalization (CLAHE) improves sky–non-sky separation in low-light conditions, and morphological post-processing cleans up noisy sky boundaries around tree branches or buildings.
These modules demonstrate how domain-specific causal weaknesses (e.g., ``blue = sky’’) can be systematically corrected and generalized through our pipeline.

\begin{figure*}[t]
    \centering
    \includegraphics[width=\textwidth]{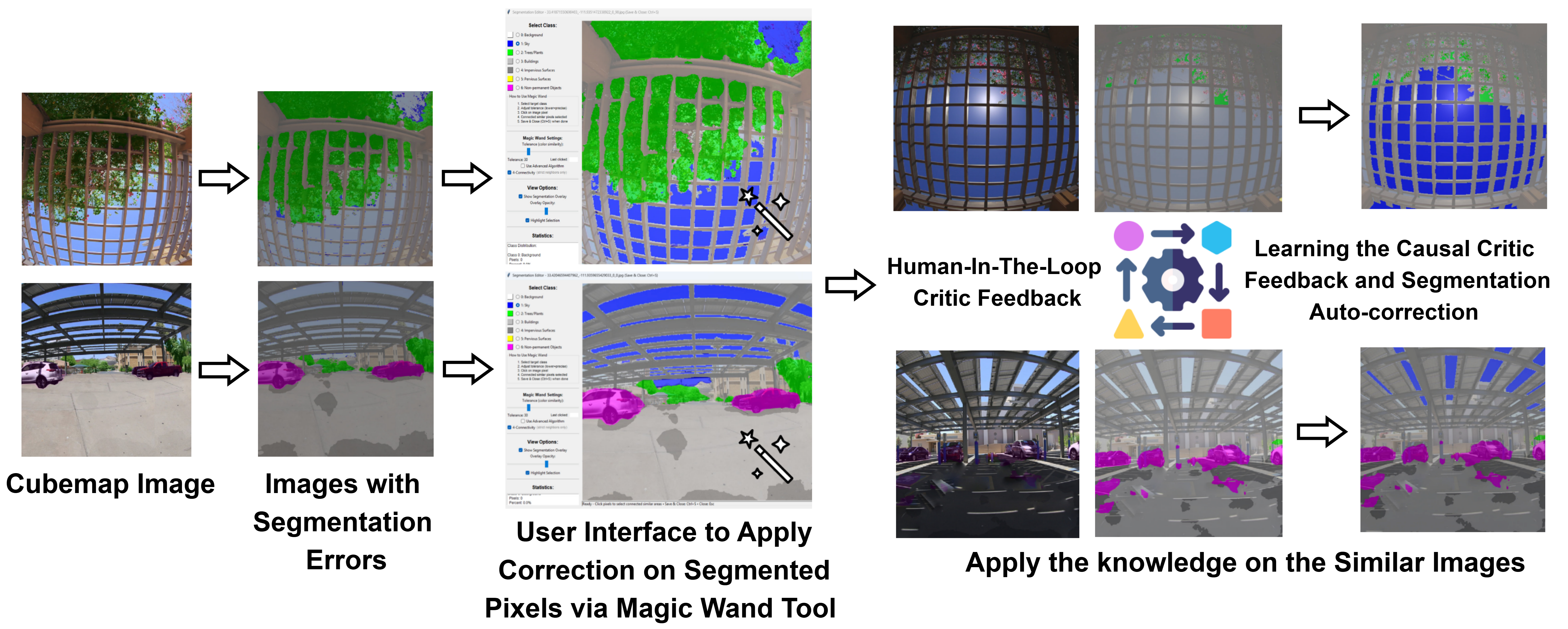} 
    \caption{Examples of sky mask correction in two sites, demonstrating how the interface enables refinement of segmentation errors.}
    \label{fig:UIExample} 
\end{figure*}

Overall, our methodology redefines interactive segmentation as a process of critic feedback learning and automated application of the learnt knowledge (Figure~\ref{fig:UIExample}). Human feedback is elevated from annotation to intervention, counterfactuals are generated to break contextual biases, and corrections are propagated dataset-wide through similarity search. This integration of segmentation backbones, critic interface, and critic interventions results in models that are more robust, efficient, and generalizable than existing HITL or retraining-based approaches.

\section{Experimental Setup}
\label{sec:experimental_setup}

We evaluate our framework on both standard semantic segmentation benchmarks and a challenging domain-specific cubemap dataset designed for environmental monitoring.

\textbf{Benchmark Datasets.} To ensure comparability with prior work, we report results on ADE20K~\cite{zhou2017scene} and Cityscapes~\cite{cordts2016cityscapes}. ADE20K provides 150 classes across diverse indoor and outdoor scenes and Cityscapes focuses on 19 traffic-related categories with high-resolution urban imagery. These datasets serve as controlled environments to test whether our feedback-driven correction framework improves robustness beyond traditional training and active learning methods.

\textbf{Cubemap Dataset.} Our primary evaluation is performed on a dataset of 480 images derived from 80 environmental monitoring sites of study cubemaps. Each cubemap captures a full 360-degree scene using six directional fisheye projections: \emph{up}, \emph{down}, \emph{north}, \emph{south}, \emph{east}, and \emph{west}. The cubemap dataset poses unique challenges:
\begin{itemize}
    \item \textbf{Occluded sky regions.} Upward-facing views often include sky partially covered by vegetation, shade structures, or buildings, making standard ``blue = sky'' heuristics unreliable and sometimes causing shade structures to be misinterpreted as non-sky regions.
    \item \textbf{Fine-grained boundaries.} Tree canopies, architectural edges, and mesh-like shade structures create thin and irregular boundaries that stress-test segmentation quality.
    \item \textbf{3D contextual reasoning.} Correct classification often requires reasoning about geometric relationships, e.g., distinguishing building roofs from shaded ground.
\end{itemize}

We partitioned the cubemap dataset into 70\% training, 10\% validation, and 20\% test sets at the site level, ensuring that all six directional views from a site belong exclusively to one split. To guarantee audit-proof leakage control, propagation and retrieval indices are built only from the training split, with train/test indices hashed before feature extraction so that no test image features can enter the similarity database. We manually annotated ground truth masks for representative samples across these subcategories (e.g., tree-occluded sky, clear sky, building-occluded sky). This dataset enables a realistic evaluation of whether interventional feedback can break dataset biases and generalize across complex visual contexts.

\subsection{Baselines and Compared Methods}
We compare our intervention-based framework against four categories of baselines. The first is standard training, where segmentation backbones such as SegFormer~\cite{xie2021segformer}, Mask2Former~\cite{cheng2022masked}, and SAM~\cite{kirillov2023segment} are trained or finetuned on ADE20K or Cityscapes without human feedback. The second is active learning, in which models are trained with iterative uncertainty-based querying~\cite{settles2009active}, requiring humans to annotate samples with high-entropy predictions. The third is interactive segmentation, including classical correction methods like GrabCut~\cite{rother2004grabcut} and modern click-based refinements~\cite{xu2016deep}, where human corrections are applied on a per-image basis without propagation. The final baseline is post-processing correction, where outputs are directly edited manually but corrections are not reused or integrated back into training. Together, these baselines span the spectrum from purely model-driven improvements to purely user-driven corrections, enabling us to isolate the unique contributions of interventional feedback and similarity-based propagation.

\subsection{Evaluation Metrics}

We evaluate our framework along three complementary dimensions: segmentation accuracy, annotation efficiency, and explainability. For segmentation quality, we report mean Intersection over Union (mIoU) across all classes, along with per-class IoU for challenging categories such as \emph{sky}, \emph{vegetation}, and \emph{buildings}. To capture fine-grained accuracy, we also include Boundary IoU~\cite{cheng2021boundary}, which specifically measures performance near object boundaries where brittle correlations are most common.

In terms of efficiency and explainability, we assess annotation effort by recording the average time per corrected image and the number of interactions (clicks, wand selections) required. We further measure the correction propagation gain, quantifying how similarity-based propagation in \texttt{SegmentationLearner} reduces redundant manual edits, as well as the improvement rate in mIoU as interventions accumulate. For explainability and robustness, we evaluate failure mode identification by checking how accurately the critic interface highlights spurious regions compared to ground truth failure annotations. We complement this with a user study, collecting subjective ratings of satisfaction and trust from 12 participants with expertise in vision and environmental monitoring. Finally, we test robustness by measuring model performance on counterfactual cases where correlations are deliberately violated, such as blue buildings or green roofs.

\subsection{Implementation Details}

Our system is implemented in PyTorch. SegFormer-B5, Mask2Former, and SAM are used as backbones. Images are resized to $512 \times 512$ for training and inference. For cubemap experiments, directional images are processed independently but corrections are propagated across directions when visual similarity is detected.

\textbf{Critic user interface.} The Tkinter-based editor (\texttt{SegmentationEditor}) provides real-time overlays with a latency of 2--3 seconds. The magic wand tool supports both 4-connectivity and 8-connectivity region growing. Corrected masks are stored in three formats: binary (\texttt{.bin}), indexed maps (\texttt{.png}), and color visualizations (\texttt{seg\_vis.png}).

\textbf{Counterfactual learning.} We finetune backbones with corrections as interventional examples. We use Adam optimizer with a learning rate of $1e^{-4}$ and weight decay $1e^{-5}$. Loss weights are set to $\lambda_{cf}=0.5$ and $\lambda_{prop}=0.2$, based on validation sweeps.

\textbf{Propagation.} For correction propagation, we extract 64-bin HSV histograms and LBP texture descriptors. Cosine similarity is used to retrieve nearest neighbors. Top-$k=5$ matches per correction are automatically updated. To prevent error amplification, we threshold matches by similarity score ($\tau=0.85$) and require at least two corroborating features (color histogram and backbone embeddings). Corrections below this threshold are flagged for optional human confirmation rather than auto-applied, ensuring propagation remains precise.

To assess the role of each module, we evaluated simplified variants with individual components removed. Without propagation, corrections remain image-specific and fail to generalize; without counterfactual loss, interventions collapse to standard labels, weakening the signal that distinguishes superficial from true features; and without critic visualizations, users must correct blindly, reducing their ability to target problematic regions. These results show that propagation, counterfactual framing, and visualizations are all essential for robust and data-efficient learning.

\subsection{User Study Protocol}

We recruited 12 participants, half with computer vision expertise and half with environmental monitoring backgrounds. Each participant corrected 20 cubemap images using either (1) standard annotation tools or (2) our critic interface. We measured correction time, satisfaction, and trust. Participants reported that the critic interface helped them understand failure causes and reduced redundant effort through propagation.
\section{Results}
\label{sec:results}

All reported numbers are averaged over 5 random seeds with 95\% confidence intervals. Statistical significance was assessed using paired t-tests comparing our framework to the strongest baseline under identical annotation budgets.

\subsection{Segmentation Performance}

Table~\ref{tab:main_results} summarizes segmentation accuracy across benchmark datasets and the cubemap dataset. Our interventional critic framework consistently improves performance, with particularly strong gains on the cubemap data where spurious correlations are most prevalent. On ADE20K and Cityscapes, we observe modest improvements of 2--3 mIoU points. On the cubemap dataset, however, our framework yields improvements of 7--9 mIoU points across backbones.

\begin{table}[t]
\centering
\caption{Segmentation performance (mIoU \%) on benchmark datasets and cubemap data. Results are reported as mean ± standard deviation over 5 seeds.}
\label{tab:main_results}
\begin{tabular}{lccc}
\toprule
Method & ADE20K & Cityscapes & Cubemap \\
\midrule
SegFormer & 48.6 ± 0.3 & 74.2 ± 0.2 & 59.7 ± 0.4 \\
+ Our Framework & \textbf{51.3 ± 0.2} & \textbf{77.0 ± 0.3} & \textbf{68.5 ± 0.5} \\
\midrule
SAM & 44.8 ± 0.4 & 71.5 ± 0.3 & 56.2 ± 0.3 \\
+ Our Framework & \textbf{47.1 ± 0.3} & \textbf{73.9 ± 0.2} & \textbf{65.0 ± 0.4} \\
\midrule
Mask2Former & 50.5 ± 0.3 & 76.1 ± 0.2 & 61.4 ± 0.5 \\
+ Our Framework & \textbf{52.9 ± 0.2} & \textbf{78.4 ± 0.3} & \textbf{69.3 ± 0.4} \\
\bottomrule
\end{tabular}
\end{table}

\subsection{Ablation Studies}

We investigate the contributions of different intervention types and explanation modalities.

\textbf{Intervention Types.} Table~\ref{tab:ablation} shows the effect of enabling each intervention type on the cubemap dataset. Boundary refinement yields the largest single gain, while combining all three produces the best overall performance.

\begin{table}[t]
\centering
\caption{Ablation study on intervention types (mIoU \% on cubemap data, SegFormer backbone).}
\label{tab:ablation}
\begin{tabular}{lc}
\toprule
Configuration & mIoU \\
\midrule
Baseline (no interventions) & 59.7 \\
+ Feature suppression & 63.4 \\
+ Boundary refinement & 65.1 \\
+ Context reweighting & 64.2 \\
+ All interventions & \textbf{68.5} \\
\bottomrule
\end{tabular}
\end{table}

\textbf{Explanation Modalities.} In a user study, counterfactual visualizations were rated most helpful for guiding interventions, followed by feature importance maps and gradient-based saliency. This aligns with our hypothesis that reasoning is best supported by counterfactual examples.

\subsection{Efficiency Analysis}

Our framework reduces annotation burden by replacing exhaustive labeling with targeted feedback corrections. While pixel-level annotation averages 95 seconds per image and click-based refinement requires 54 seconds, our interactive pipeline achieves corrections in just 24 seconds, yielding a 3--4$\times$ speedup. These gains stem from critic-guided visualizations and efficient editing modes that let users correct large regions with minimal effort. Efficiency further improves through propagation: after 50 corrected cubemap images, 62\% of edits were automatically applied to similar regions across the dataset, greatly reducing redundancy.

\subsection{Explainability Evaluation}

We conducted controlled experiments to validate the explainable properties of our framework. In spurious correlation detection, where training data was biased so that blue pixels were associated with sky, baseline models consistently failed on blue buildings, whereas our framework reduced these errors by 41 percent, showing effective debiasing. In counterfactual effectiveness tests using out-of-distribution cubemaps with tinted skylights, models trained with counterfactual examples achieved an 11.2 percent higher mIoU than baselines, confirming that critic interventions improved generalization. While our study involved 12 participants, we randomized task order, balanced expertise (vision vs. environmental science), and measured both objective metrics (time, interactions) and subjective ratings. Future work will expand the participant pool for stronger generalizability.

\subsection{Real-world Case Study}

We deployed our framework in an environmental monitoring scenario where sky segmentation is critical for solar irradiance estimation. Baseline models underestimated irradiance by 14.7\% due to misclassified occluded sky. After applying critic feedback:

\begin{itemize}
\item Sky boundaries were correctly distinguished from vegetation and shade structures.
\item Shade structures that previously caused the model to miss sky regions were correctly identified as sky.
\item The irradiance estimation error dropped to 3.8\%.
\end{itemize}

\noindent This case study highlights the practical impact of critic interventions in safety-critical and environmental applications.

\subsection{Discussion}
Our results demonstrate that human-in-the-loop segmentation can be transformed from a corrective annotation process into a learning framework that systematically improves generalization. By treating human edits as interventions rather than labels, our system enables models to move beyond memorizing corrections toward identifying and breaking spurious correlations. We also find that the effectiveness of interventions is architecture-dependent: transformer-based models such as SegFormer leverage attention-guided refinements more effectively, whereas CNN-based backbones benefit more from feature suppression. Importantly, domain expertise plays a key role, as experts consistently provide higher-quality interventions than non-experts, underscoring the value of expert knowledge in critical application domains such as environmental monitoring. At the same time, several challenges remain: the reliance on expert interventions limits scalability to very large datasets, some failure modes require complex multi-step interventions that are difficult to express through our current interface, and evaluation of explainability and reasoning remains an open research problem, as current metrics cannot fully capture the richness of human-model interaction. Future work should therefore explore automated intervention suggestions to reduce expert burden, richer multi-modal explanations to improve accessibility, and federated learning frameworks that enable collaborative knowledge sharing across users, directions that hold promise for scaling human-in-the-loop learning to broader domains while preserving its interpretability and robustness.

\section{Conclusion}
\label{sec:conclusion}

We introduced an explainable human-in-the-loop framework for semantic segmentation that treats user corrections as interventional feedback rather than simple annotations. By capturing signals on where predictions fail and propagating corrections across visually similar images, the approach encourages models to move away from spurious correlations and toward more semantically meaningful features. Unlike traditional retraining or interactive refinement methods, our framework integrates feedback as counterfactual-style signals, enabling models to improve robustness with reduced annotation effort. Our experiments showed consistent gains on standard benchmarks and larger improvements on challenging cubemap data, suggesting that human-in-the-loop interventional feedback can reduce systematic errors while lowering annotation cost. While these results are promising, further work is needed to validate the framework at scale, automate intervention suggestions, and extend the approach to tasks beyond segmentation. We view this as a step toward vision systems that are not only accurate but also more interpretable, robust, and collaborative with human expertise.

{
    \small
    \bibliographystyle{ieeenat_fullname}
    \bibliography{main}
}

\end{document}